\documentclass[11pt]{article}

\usepackage[preprint]{acl}

\usepackage{times}
\usepackage{latexsym}
\usepackage[T1]{fontenc}
\usepackage[utf8]{inputenc}
\usepackage{microtype}
\usepackage{inconsolata}
\usepackage{graphicx}
\usepackage{amsmath}
\usepackage{amssymb}
\usepackage{booktabs}
\usepackage{multirow}
\usepackage{tabularx}
\usepackage{colortbl}
\usepackage{arydshln}
\usepackage{pifont}
\newcommand{\cmark}{\ding{51}}
\newcommand{\xmark}{\ding{55}}
\definecolor{teal}{rgb}{0,0.5,0.5}

\title{EvoAgentBench: Benchmarking Agent Self-Evolution via Ability Transfer}

\author{%
  Xingze Gao\textsuperscript{1,2}\quad
  Chuanrui Hu\textsuperscript{2\,\dag}\quad
  Hongda Chen\textsuperscript{2,3}\quad
  Pengfei Yao\textsuperscript{2}\quad
  Zhao Wang\textsuperscript{2}\quad
  Yi Bai\textsuperscript{2}\\
  \textbf{Zhengwei Wu\textsuperscript{2}\quad
  Yunyun Han\textsuperscript{2}\quad
  Xiaofeng Cong\textsuperscript{3}\quad
  Jie Gui\textsuperscript{3}\quad
  Yafeng Deng\textsuperscript{2\,*}\quad
  Teng Li\textsuperscript{1\,*}}\\[4pt]
  \textsuperscript{1}Anhui University\quad
  \textsuperscript{2}EverMind, Shanda Group\quad
  \textsuperscript{3}Southeast University\\
  \texttt{\{xingze.gao, chuanrui.hu, hongda.chen, yaopengfei\}@shanda.com}\\
  \texttt{\{zhao.wang, baiyi, hanyunyun, dengyafeng\}@shanda.com}\\
  \texttt{zhengwei.wu@evermind.ai\quad tenglwy@gmail.com}\\
  \texttt{cxf\_svip@163.com\quad guijie@ustc.edu}%
}

\begin{document}
\maketitle
\renewcommand*{\thefootnote}{\fnsymbol{footnote}}
\footnotetext[1]{Co-corresponding authors.}
\footnotetext[2]{Project leader.}
\renewcommand*{\thefootnote}{\arabic{footnote}}

\begin{abstract}
Agent self-evolution in long-horizon LLM systems is largely procedural: useful experience is not merely stored information, but reusable procedures for searching, debugging, and verification. Yet current evaluations do not isolate this form of transfer. Agent benchmarks test single-episode task solving; memory benchmarks target information retention rather than procedural reuse. We introduce \textbf{EvoAgentBench}, a benchmark for agent self-evolution via Ability-guided transfer across four agentic domains: web research, algorithmic reasoning, software engineering, and knowledge work. EvoAgentBench extracts trace-grounded Abilities from agent executions, canonicalizes them into operational units, and builds domain-specific Ability Graphs linking tasks that share procedural overlap. By design, every test task is backed by verified training-side Ability support. Across a 528/267 train/test split, two scaffolds, and three backbones, curated Ability content transfers reliably across model families, but no current automatic method sustains positive gain in all settings. EvoAgentBench shifts self-evolution evaluation from aggregate accuracy comparison to fine-grained diagnosis of experience encoding, routing, and uptake. The benchmark is publicly available at \url{https://huggingface.co/datasets/EverMind-AI/EvoAgentBench}.
\end{abstract}

\section{Introduction}

A useful agent should not solve every task from scratch. After prior executions,
it should retain reusable procedures for searching, debugging, and verification.
As LLM agents increasingly master isolated long-horizon
tasks~\citep{patwardhan2025gdpval,chen2025browsecomp,jain2024livecodebench,
jimenez2024swebench}, the evaluation frontier is shifting toward exactly this
capability: improvement from experience.

We study skill-based agent self-evolution, where past trajectories are distilled
into external artifacts such as skills, cases, workflows, or evolved prompts, and
made available when solving future tasks. This paradigm offers a lightweight, model-agnostic path to improving agent
reliability~\citep{shinn2023reflexion,zhao2024expel,wang2025agent,wangvoyager,
fernando2024promptbreeder}.

Current evaluations fail to isolate trajectory-to-procedure transfer. Task-centric
benchmarks measure whether an agent can solve a held-out task, but they never test
whether prior experience converts into reusable procedural
knowledge~\cite{patwardhan2025gdpval,chen2025browsecomp,jain2024livecodebench,
jimenez2024swebench,mialon2024gaia}. Long-context and memory benchmarks evaluate
recall, retrieval, and personalization over stored information, but stop at
information retention, never reaching procedural
reuse~\cite{bai2024longbench,maharana2024evaluating,wulongmemeval}.

Benchmarks for self-evolving agents come closer but still fall short. These
evaluate how skill libraries grow over sequential tasks, yet they entangle
artifact quality with task order, retrieval policy, and scaffold
dynamics~\cite{cai2025stulife,wu2024streambench,jiang2026seaeval}.
Skill-oriented benchmarks probe procedure quality, but evaluation operates within
individual tasks rather than measuring cross-task
transfer~\cite{li2026skillsbench,zhong2026skilllearnbench}. The result is a
shared blind spot: \textbf{no current benchmark explicitly controls for
ability-level support while measuring whether reusable procedures transfer to
unseen tasks with related but non-identical demands.}

Evaluating trajectory-to-procedure transfer introduces a tension absent from
ordinary held-out evaluation. Evaluation tasks must be related enough to prior
experience that transferable procedures exist, yet distinct enough that
improvement cannot stem from memorized answers or near-duplicate exposure.
\textit{How can a benchmark guarantee both?} We resolve this with a single
structural principle: a benchmark for agent self-evolution must be
ability-supported yet instance-disjoint.

We introduce \textbf{EvoAgentBench}, a multi-domain benchmark built around this
principle. It provides training trajectories from which self-evolution methods
derive artifacts, together with held-out tasks requiring related capabilities
without duplicating instances. The benchmark spans four long-horizon agentic
domains: web research, algorithmic reasoning, software engineering, and
knowledge work. Rather than using benchmark or domain labels as transfer units,
EvoAgentBench extracts trace-grounded Abilities from agent executions to build
domain-specific Ability Graphs whose splits measure procedure-level transfer
rather than random task separation. Under matched external conditions, automatic methods
differ only in how they encode and reuse training experience; the benchmark's
own Ability-grounded skills serve as a diagnostic reference with curator-side routing.

Using EvoAgentBench, we expose a substantial gap between high-quality procedural
content and current automatic self-evolution methods. The Ability-grounded
reference condition improves held-out performance across all four domains, but
automatic methods remain brittle, with gains varying across settings. We also
identify instances of negative transfer, where injected artifacts hurt task
performance. Finally, interaction cost is an unreliable proxy for accuracy,
motivating joint reporting of performance and overhead.

In summary, this paper makes three contributions:
\begin{itemize}
  \item \textbf{Evaluation framework.} We formulate agent self-evolution as a
    controlled experience-to-transfer evaluation problem and provide a matched
    evaluation substrate so that methods differ only in how they encode and reuse
    training experience.
  \item \textbf{Benchmark.} We introduce EvoAgentBench, a multi-domain benchmark
    whose trace-grounded Ability Graphs enable procedure-level transfer measurement
    across four agentic domains; the benchmark's own Ability-grounded skills serve
    as a diagnostic reference.
  \item \textbf{Empirical findings.} Evaluation across two scaffolds and multiple
    backbones shows that reusable procedural content transfers when correctly
    delivered, but current automatic methods remain brittle across domains,
    scaffolds, and cost regimes.
\end{itemize}

\section{Related Work}
\begin{table*}[t]
  \centering
  \footnotesize
  \setlength{\tabcolsep}{4.5pt}
  \renewcommand{\arraystretch}{1.1}
  \begin{tabular}{@{}l l cc c r@{}}
    \toprule
    \textbf{Benchmark}
      & \textbf{Eval Target}
      & \textbf{Cross-Task}
      & \textbf{Abil.-Aware Split}
      & \textbf{\#D}
      & \textbf{\#Tasks} \\
    \midrule
    LongMemEval            & Info.\ retention   & \xmark & \xmark & 1  & 500 \\
    AMA-Bench              & Traj.\ memory      & \xmark & \xmark & 6  & 3{,}696 \\
    \midrule
    StreamBench            & Online adapt.       & \cmark & \xmark & 5  & 9{,}702 \\
    LifelongAgentBench     & Online adapt.       & \cmark & \xmark & 3  & 1{,}396 \\
    SEA-Eval               & Online adapt.       & \cmark & \xmark & 2  & 90 \\
    \midrule
    SkillsBench            & Skill quality       & \xmark & \xmark & 11 & 84 \\
    SkillLearnBench        & Skill learning      & \xmark & \xmark & 6  & 100 \\
    SkillFlow              & Skill evolution     & \cmark & \xmark & 5  & 166 \\
    \midrule
    \textbf{EvoAgentBench} & Proc.\ transfer    & \cmark & \cmark & 4  & 528\,/\,267$^\star$ \\
    \bottomrule
  \end{tabular}
  \caption{Comparison with related benchmarks. \#D: number of domains. $^\star$\,Train\,/\,test split; other benchmarks report undivided evaluation pools or sequential streams.}
  \label{tab:benchmark-comparison}
\end{table*}

\paragraph{Agent self-evolution via reusable artifacts.}
Methods for non-parametric self-evolution can be organized by artifact type.
Reflexion \citep{shinn2023reflexion} introduced verbal reinforcement via transient reflections;
ExpeL \citep{zhao2024expel} generalized this across tasks into natural-language insights;
Voyager \citep{wangvoyager} built executable skill libraries.
AWM \citep{wang2025agent} targets workflows, while prompt-evolution systems such as Promptbreeder and GEPA \citep{fernando2024promptbreeder,agrawal2025gepa} target evolved prompts.
Continuously updated memories include Dynamic Cheatsheet \citep{suzgun2025dynamiccheatsheet} and streaming reasoning pools \citep{xiao2025reasoningbank},
while Memento \citep{xu2025memento} retains per-task cases under a learned retrieval policy.
A recent wave converges on curated skill packages that distill trajectories into structured, persistent files~\citep{zhou2026mementoskills,ni2026trace2skill,ouyang2026skillos,mi2026procmem,yang2026autoskill}.
Yet no method explicitly models which reusable abilities are present in training and whether they transfer. The gap is not artifact diversity but transfer awareness: without modeling which abilities are present at training time, a method's improvement could stem from its extracted procedures, retrieval, or incidental task-level overlap.
EvoAgentBench grounds transfer evaluation in trace-derived Ability units and compares automatic self-evolution methods against a diagnostic reference under matched conditions.

\paragraph{Benchmarks for self-evolving agents.}
Existing benchmarks are largely streaming:
LifelongAgentBench \citep{zheng2025lifelongagentbench}, SEA-Eval \citep{jiang2026seaeval}, StuLife \citep{cai2025stulife}, and Evo-Memory \citep{wei2025evomemory} study how memory or skill libraries evolve as tasks arrive sequentially, conflating artifact quality with scheduling and retrieval dynamics.
AMA-Bench \citep{amabench2026} targets in-episode agentic memory rather than cross-task procedural transfer.
SkillsBench \citep{li2026skillsbench}, SkillFlow \citep{zhang2026skillflow}, and SkillLearnBench \citep{zhong2026skilllearnbench} probe skill-artifact quality: SkillsBench reports that curated skills substantially outperform self-generated ones, and even SkillLearnBench's three-level evaluation operates within individual tasks.
Yet when a method fails, none can pinpoint whether the bottleneck lies in missing training-side ability, flawed extraction, or retrieval breakdown, a distinction essential to method development.

EvoAgentBench closes this gap (Table~\ref{tab:benchmark-comparison}) with an Ability Graph that guarantees training-side Ability support for every test task, enabling diagnosable transfer evaluation.

\section{EvoAgentBench}
\label{sec:evoagentbench}

\begin{figure*}[t]
\centering
\includegraphics[width=\textwidth]{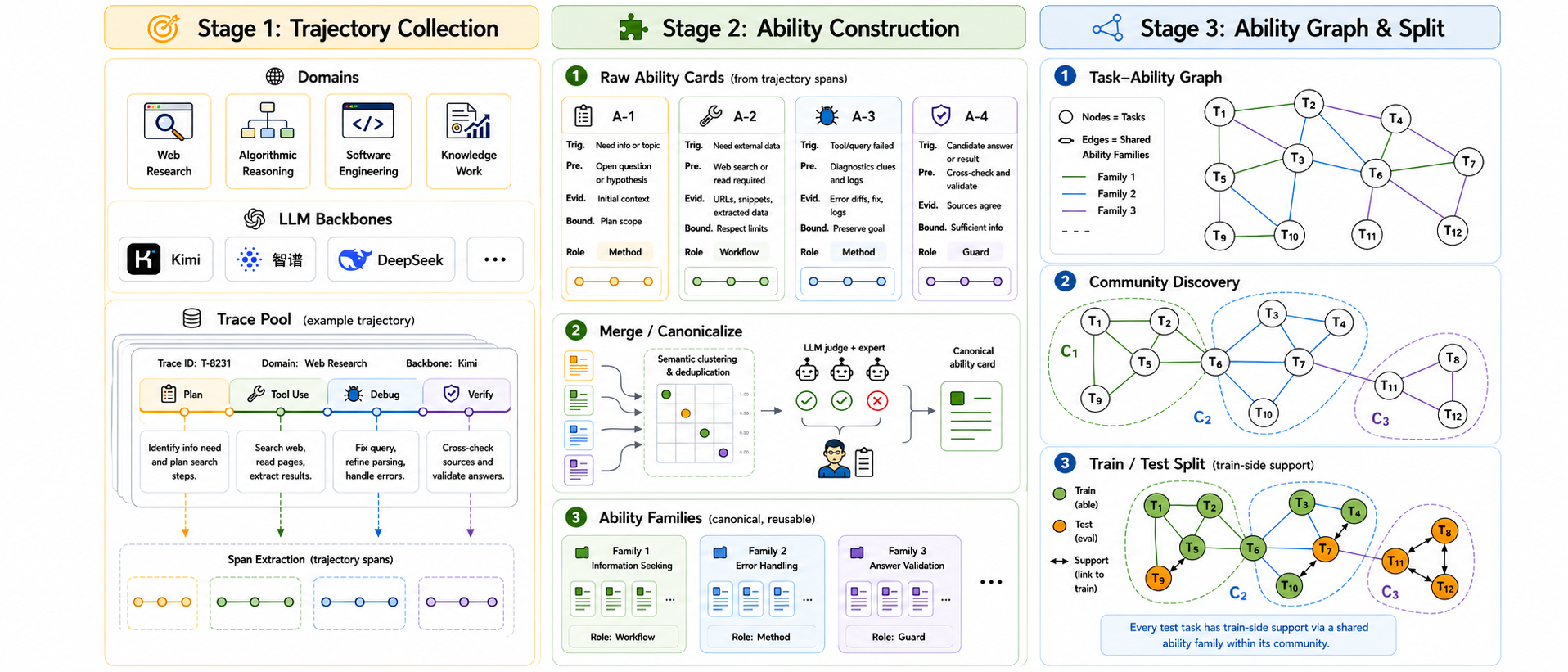}
\caption{
EvoAgentBench construction pipeline. \textbf{Stage~1}: no-skill executions from multiple backbones across four domains yield a trace pool with reusable spans. \textbf{Stage~2}: raw Ability cards are merged into canonical families (Method, Guard, or Workflow) via embedding blocking, LLM adjudication, and expert review. \textbf{Stage~3}: the task--Ability graph is partitioned into communities, and constrained splitting guarantees train-side Ability support for every test task.
}
\label{fig:protocol}
\end{figure*}

EvoAgentBench is built around an \emph{Ability Graph}. Nodes are tasks, and edges indicate shared reusable Abilities extracted from agent executions. An Ability is not a dataset label, topic, or repository name; it is a reusable operation such as a search strategy, debugging procedure, or validation workflow. Task relatedness, data splitting, and diagnostic construction all derive from these operations rather than surface metadata. Figure~\ref{fig:protocol} overviews the three-stage construction pipeline.

\subsection{Benchmark Formulation}
\label{sec:benchmark-formulation}

Let $D_{\mathrm{train}}$ and $D_{\mathrm{test}}$ denote the training and test splits. A self-evolution method $m$ receives training-side evidence only---task prompts, verifier outcomes, associated metadata, and training trajectories, either the benchmark's construction traces or rollouts the method itself performs on $D_{\mathrm{train}}$ (the protocol in our experiments; Section~\ref{sec:experiments})---and produces an evolution state $z_m$ encoding this experience into a reusable form such as a skill library, case bank, workflow set, or evolved prompt. The benchmark is not tied to a particular representation; future methods may instantiate $z_m$ as lightweight updates, provided the update derives only from training-side evidence.

During evaluation, the agent is run on $D_{\mathrm{test}}$ with $z_m$ available through the method's own interface. Let $r_0(x)$ be the no-evolution baseline score on test task $x$, and $r_m(x)$ the score after applying method $m$. The primary observable is the average transfer gain:
\[
\Delta_m =
\frac{1}{|D_{\mathrm{test}}|}
\sum_{x\in D_{\mathrm{test}}}
\left(r_m(x)-r_0(x)\right).
\]
We additionally report cost changes when available.

Within each agent--backbone setting, we fix the task statement, tool set, scoring contract, timeout, and base agent configuration. Methods differ only in what evolution state they construct and how that state is applied at test time, and EvoAgentBench measures whether each chosen form of self-evolution improves test-time behavior on tasks with verified train-side Ability support.

\subsection{Ability-Guided Data Construction}
\label{sec:ability-guided-construction}

EvoAgentBench covers four long-horizon agentic domains: web research, algorithmic reasoning, software engineering, and knowledge work, instantiated with BrowseComp-Plus, LiveCodeBench, SWE-Bench Verified, and GDPVal respectively~\citep{chen2025browsecomp,jain2024livecodebench,jimenez2024swebench,patwardhan2025gdpval}. Each domain favors distinct reusable procedures: search and verification strategies for web research, algorithm design and debugging patterns for algorithmic reasoning, repository-level repair procedures for software engineering, and artifact construction or validation workflows for knowledge work. Rather than treating source benchmark labels or domain metadata as transfer units, EvoAgentBench builds transfer structure from \textbf{trace-grounded Abilities}.

\subsubsection{Multi-Backbone Trace Collection}

Ability construction is grounded in agent behavior rather than task metadata alone. For each task, we collect executions from a set of construction backbones $\mathcal{B}_{\mathrm{trace}}$. In our implementation, $\mathcal{B}_{\mathrm{trace}}$ contains three construction backbones: Kimi-K2.5, GLM-5.1, and DeepSeek-V3.2~\citep{kimiteam2026kimik25visualagentic,glm5team2026glm5vibecodingagentic,deepseekai2025deepseekv32pushingfrontieropen}.

For each task $x$ and backbone $b\in\mathcal{B}_{\mathrm{trace}}$, we collect executions across construction scaffolds (Appendix~\ref{app:trace-collection}); $\tau_{x,b}$ denotes the resulting trace set, with $o_{x,b}$ and $\nu_{x,b}$ the corresponding outputs and verifier scores (binary, scalar, or structured depending on the source benchmark). The task-level multi-backbone trace pool is $\mathcal{T}(x)=\{(\tau_{x,b}, o_{x,b}, \nu_{x,b}) : b\in\mathcal{B}_{\mathrm{trace}}\}$.
The extractor compares successful and failed attempts across backbones, identifying recurring procedures, failure modes, and discriminative verification steps at the task level. This cross-backbone aggregation reduces sensitivity to any single model's shortcuts, tool-use habits, or idiosyncratic failures.

\subsubsection{Trace-Grounded Ability Extraction}
\label{sec:trace-grounded-ability-extraction}

EvoAgentBench defines transfer over \emph{Abilities}: trace-grounded abstractions of reusable operations with explicit applicability conditions, specifying when an agent should apply a procedure, correction, or validation step.

For each task $x$, the extractor observes the task query $q_x$, the ground-truth target $y_x^\star$, and the multi-backbone trace pool $\mathcal{T}(x)$, and produces task-local raw Ability cards $\mathcal{A}_{\mathrm{raw}}(x)=E_{\phi}\bigl(q_x, y_x^\star, \mathcal{T}(x)\bigr)$.

Each raw Ability card is a tuple $a=(\gamma_a,\pi_a,\mathcal{E}_a,\partial_a,\rho_a)$, where $\gamma_a$ is a trigger condition, $\pi_a$ is the reusable procedure, and $\mathcal{E}_a$ is supporting evidence linking the card to trajectory spans, outputs, verifier results, and the ground-truth target. The boundary $\partial_a$ restricts when the procedure should transfer, preventing broad instructions from forming spurious Ability links, and $\rho_a$ assigns one of three roles: \textbf{Method} for primary task-solving procedures, \textbf{Guard} for corrections against recurring invalid behavior, and \textbf{Workflow} for execution-control procedures such as verification. These roles support analysis and auditing but do not define separate evaluation metrics.

The extractor operates jointly over multi-backbone evidence for each task rather than processing individual runs independently, letting $E_{\phi}$ compare successful and failed executions, identify operations that distinguish reliable completions from brittle attempts, and abstract recurring failure patterns into corrective procedures.

\subsubsection{Ability Canonicalization}

The raw Ability cards produced by $E_\phi$ are task-local: different tasks may describe the same operation with different wording, while superficially similar cards may correspond to different procedures. EvoAgentBench therefore canonicalizes raw cards through conservative adjudication targeting operational equivalence rather than semantic clustering.

For each domain $d$, we use embedding similarity as a recall-oriented blocking step over the domain's raw cards $\mathcal{A}_{\mathrm{raw}}^d=\bigcup_{x\in D_d}\mathcal{A}_{\mathrm{raw}}(x)$. We embed each card using its trigger, procedure, boundary, and role fields, and take as candidates all pairs whose embedding cosine similarity meets a domain-specific threshold $\theta_d$. The threshold serves only to select pairs for adjudication; embedding similarity is never a merge criterion.

Each candidate pair is judged under an operational-equivalence rubric. A merge requires the same role type, compatible triggers, equivalent reusable procedures, the same success mechanism or correction target, and compatible applicability boundaries. Shared topic, lexical overlap, or generic verbs (``search'', ``debug'', ``validate'') are insufficient.

To reduce single-judge variance, each pair is evaluated by three independent LLM adjudicators; unanimously approved pairs are accepted automatically, and all others are reviewed by domain experts under the same rubric, yielding pair-level merge and cannot-link decisions.

Merge decisions are not transitively closed: $a_i\!\sim\!a_j$ and $a_j\!\sim\!a_k$ does not imply that $a_i$, $a_j$, $a_k$ form one Ability. Accepted links define a compatibility graph, but a connected component becomes a canonical Ability unit only after a group-level consistency check confirms that every internal pair is merge-compatible and none carries a cannot-link decision. Domain experts split components that violate this condition, assigning broad bridge cards to their most specific compatible unit or downgrading them to annotation-only status.

Each accepted unit $u=(\mathcal{R}_u,\Gamma_u,\Pi_u,\mathcal{E}_u,\partial_u,\rho_u,\mathcal{X}_u)$ aggregates the member raw cards $\mathcal{R}_u$ into canonical fields ($\Gamma_u$ trigger, $\Pi_u$ procedure, $\mathcal{E}_u$ aggregated evidence, $\partial_u$ boundary, $\rho_u$ role), with supporting task set $\mathcal{X}_u=\{x:\mathcal{R}_u \cap \mathcal{A}_{\mathrm{raw}}(x)\neq\varnothing\}$.
Units whose procedure becomes too generic after merging are retained for annotation rather than used as transfer-defining Abilities. The resulting canonical units form the shared Ability vocabulary for constructing the Ability Graph.

\subsubsection{Ability Graph Construction}
\label{sec:ability-graph-construction}

We map canonical units back to tasks through their supporting raw cards. Let $\mathcal{U}_d$ denote the canonical Ability units in domain $d$; the Abilities assigned to task $x$ are $A(x)=\{u\in\mathcal{U}_d:x\in\mathcal{X}_u\}$.

Not every canonical unit defines a transfer edge. A unit is \emph{edge-eligible} if at least two distinct tasks support it and its procedure remains operationally specific after merging; singleton, overly broad, and annotation-only units are retained for audit but cannot create graph edges. Let $A^+(x)\subseteq A(x)$ denote the edge-eligible Abilities of task $x$.

For each domain $d$, we construct an undirected Ability Graph $G_d=(V_d,E_d)$, where $V_d$ contains the source tasks and an edge $(x_i,x_j)\in E_d$ is added exactly when two tasks share at least one edge-eligible Ability, i.e., $A^+(x_i)\cap A^+(x_j)\neq\varnothing$.
Each edge stores the shared Ability identifiers and their roles, enabling later analysis of whether a relation is induced by Method, Guard, or Workflow Abilities. Tasks with no edge-eligible overlap remain isolated and are assigned to the training set, since they cannot satisfy the test-split support constraint.

We then partition each domain graph into Ability communities via Louvain community detection: local regions of procedural overlap, not external labels, used only to localize train-side support during split construction.
\subsubsection{Ability-Aware Transfer Split}
\label{sec:ability-aware-transfer-split}

We restrict to the transferable pool $D_d^{\mathrm{pool}} = \{x \in D_d : A^+(x) \neq \varnothing\}$ of tasks with at least one edge-eligible Ability, and sample the evaluation split within each domain's Ability Graph by community, so that train-side support comes from the same local Ability neighborhood: every test task must share at least one edge-eligible Ability with a training task in its own community.
We perform constrained sampling under this invariant: moving a task into the test split is allowed only if it and all previously selected test tasks retain train-side support. Among feasible assignments, no-evolution verifier scores serve as a soft headroom signal, so that the test split is not dominated by tasks already solved or failed by all construction backbones.

The resulting evaluation split contains $528$ training tasks and $267$ test tasks, with zero unsupported test tasks.

\subsection{Dataset Statistics and Construction Checks}
\label{sec:dataset-statistics}

\begin{table}[t]
  \centering
  \footnotesize
  \setlength{\tabcolsep}{3pt}
  \renewcommand{\arraystretch}{1.08}
  \begin{tabular}{@{}l rr rrr@{}}
    \toprule
    & \multicolumn{2}{c}{\textbf{Eval.\ Tasks}} & \multicolumn{3}{c}{\textbf{Full Ability Graph}} \\
    \cmidrule(lr){2-3} \cmidrule(l){4-6}
    \textbf{Domain}
      & \textbf{Train} & \textbf{Test}
      & \textbf{Abl.} & \textbf{Comm.} & \textbf{Abl./T} \\
    \midrule
    BrowseComp-Plus    & 154 &  65 & 38 & 13 & 2.42 \\
    SWE-Bench Verified &  87 &  56 & 38 & 15 & 2.21 \\
    LiveCodeBench      & 182 &  86 & 58 & 22 & 2.12 \\
    GDPVal             & 105 &  60 & 36 &  6 & 2.63 \\
    \midrule
    \textbf{Total}     & \textbf{528} & \textbf{267} & \textbf{170} & \textbf{56} & \textbf{2.36} \\
    \bottomrule
  \end{tabular}
  \caption{Dataset statistics. Train/Test report the supported evaluation split used in experiments. The Ability Graph columns summarize the full retained graph before evaluation-subset sampling. Abl.: canonical Ability units; Comm.: Ability communities; Abl./T: average Ability units per retained graph task. Zero test tasks are unsupported.}
  \label{tab:dataset-statistics}
\end{table}

Table~\ref{tab:dataset-statistics} separates the two task counts used by EvoAgentBench. The source pool contains 2{,}605 tasks across four domains. Trace-grounded extraction produces 7{,}326 raw Ability cards from 2{,}516 tasks, and adjudicated canonicalization merges these into 170 canonical Ability units. After excluding tasks with singleton, weak, or annotation-only Abilities, the full retained Ability graph contains 1{,}108 tasks. The main experiments use a supported evaluation subset of this graph, containing 528 training tasks and 267 test tasks with zero unsupported test tasks.




\begin{table*}[!t]
\centering
\fontsize{8pt}{10pt}\selectfont
\setlength{\tabcolsep}{3.5pt}
\renewcommand{\arraystretch}{1.15}

\resizebox{\textwidth}{!}{%
\begin{tabular}{l *{4}{c} @{\hspace{8pt}} *{4}{c} @{\hspace{4pt}} c}
\toprule
\multirow{2}{*}{\textbf{Method}}
& \multicolumn{4}{c}{\textbf{OpenClaw}}
& \multicolumn{4}{c}{\textbf{Nanobot}}
& \multirow{2}{*}{\textbf{Average}} \\
\cmidrule(lr){2-5}\cmidrule(lr){6-9}
& \textit{Web} & \textit{Algo} & \textit{SWE} & \textit{KW}
& \textit{Web} & \textit{Algo} & \textit{SWE} & \textit{KW} & \\
\midrule
\multicolumn{10}{c}{\cellcolor{gray!10}\textbf{Qwen3.5-27B}} \\
\midrule
Vanilla & 10.8{\scriptsize$\pm$3.4} & 51.9{\scriptsize$\pm$5.2} & 42.3{\scriptsize$\pm$5.3} & 43.6{\scriptsize$\pm$4.6} & 8.7{\scriptsize$\pm$2.6} & 52.7{\scriptsize$\pm$5.0} & 45.8{\scriptsize$\pm$5.5} & 43.9{\scriptsize$\pm$4.7} & 37.5 \\
\cdashline{1-10}
\addlinespace[2pt]
\hspace{1em} + Memento & 11.3{\scriptsize$\pm$3.0} {\scriptsize\textcolor{teal}{+0.5}} & 53.9{\scriptsize$\pm$5.0} {\scriptsize\textcolor{teal}{+2.0}} & 45.2{\scriptsize$\pm$5.4} {\scriptsize\textcolor{teal}{+2.9}} & 44.3{\scriptsize$\pm$4.3} {\scriptsize\textcolor{teal}{+0.7}} & 12.3{\scriptsize$\pm$3.1} {\scriptsize\textcolor{teal}{+3.6}} & 57.0{\scriptsize$\pm$5.2} {\scriptsize\textcolor{teal}{+4.3}} & 9.5{\scriptsize$\pm$3.4} {\scriptsize\textcolor{red}{\textminus 36.3}} & 47.6{\scriptsize$\pm$4.8} {\scriptsize\textcolor{teal}{+3.7}} & 35.1 {\scriptsize\textcolor{red}{(\textminus 2.4)}} \\
\hspace{1em} + RB & 12.8{\scriptsize$\pm$2.8} {\scriptsize\textcolor{teal}{+2.0}} & 57.4{\scriptsize$\pm$5.1} {\scriptsize\textcolor{teal}{+5.5}} & 36.9{\scriptsize$\pm$5.1} {\scriptsize\textcolor{red}{\textminus 5.4}} & 48.2{\scriptsize$\pm$4.5} {\scriptsize\textcolor{teal}{+4.6}} & 9.2{\scriptsize$\pm$3.2} {\scriptsize\textcolor{teal}{+0.5}} & 62.4{\scriptsize$\pm$5.1} {\scriptsize\textcolor{teal}{+9.7}} & 53.0{\scriptsize$\pm$5.6} {\scriptsize\textcolor{teal}{+7.2}} & 49.2{\scriptsize$\pm$4.7} {\scriptsize\textcolor{teal}{+5.3}} & 41.1 {\scriptsize\textcolor{teal}{(+3.6)}} \\
\hspace{1em} + GEPA & 14.9{\scriptsize$\pm$3.6} {\scriptsize\textcolor{teal}{+4.1}} & 53.9{\scriptsize$\pm$5.0} {\scriptsize\textcolor{teal}{+2.0}} & 41.7{\scriptsize$\pm$5.8} {\scriptsize\textcolor{red}{\textminus 0.6}} & 31.3{\scriptsize$\pm$4.0} {\scriptsize\textcolor{red}{\textminus 12.3}} & 11.8{\scriptsize$\pm$3.4} {\scriptsize\textcolor{teal}{+3.1}} & 60.5{\scriptsize$\pm$5.0} {\scriptsize\textcolor{teal}{+7.8}} & 48.8{\scriptsize$\pm$5.7} {\scriptsize\textcolor{teal}{+3.0}} & 46.6{\scriptsize$\pm$4.6} {\scriptsize\textcolor{teal}{+2.7}} & 38.7 {\scriptsize\textcolor{teal}{(+1.2)}} \\
\cdashline{1-10}
\addlinespace[2pt]
\hspace{1em} + Anchor$^\dagger$ & 14.9{\scriptsize$\pm$2.9} {\scriptsize\textcolor{teal}{+4.1}} & 60.5{\scriptsize$\pm$4.8} {\scriptsize\textcolor{teal}{+8.6}} & 52.4{\scriptsize$\pm$5.2} {\scriptsize\textcolor{teal}{+10.1}} & 48.6{\scriptsize$\pm$5.9} {\scriptsize\textcolor{teal}{+5.0}} & 13.3{\scriptsize$\pm$2.9} {\scriptsize\textcolor{teal}{+4.6}} & 63.2{\scriptsize$\pm$5.1} {\scriptsize\textcolor{teal}{+10.5}} & 49.4{\scriptsize$\pm$5.4} {\scriptsize\textcolor{teal}{+3.6}} & 57.7{\scriptsize$\pm$5.2} {\scriptsize\textcolor{teal}{+13.8}} & 45.0 {\scriptsize\textcolor{teal}{(+7.5)}} \\
\midrule
\multicolumn{10}{c}{\cellcolor{gray!10}\textbf{Qwen3.5-397B}} \\
\midrule
Vanilla & 31.8{\scriptsize$\pm$4.6} & 52.7{\scriptsize$\pm$5.2} & 66.7{\scriptsize$\pm$5.7} & 55.8{\scriptsize$\pm$4.8} & 12.8{\scriptsize$\pm$4.3} & 57.8{\scriptsize$\pm$5.1} & 62.5{\scriptsize$\pm$5.8} & 53.4{\scriptsize$\pm$4.5} & 49.2 \\
\cdashline{1-10}
\addlinespace[2pt]
\hspace{1em} + Memento & 33.8{\scriptsize$\pm$4.8} {\scriptsize\textcolor{teal}{+2.0}} & 60.9{\scriptsize$\pm$5.1} {\scriptsize\textcolor{teal}{+8.2}} & 63.7{\scriptsize$\pm$5.3} {\scriptsize\textcolor{red}{\textminus 3.0}} & 50.0{\scriptsize$\pm$4.6} {\scriptsize\textcolor{red}{\textminus 5.8}} & 16.9{\scriptsize$\pm$3.4} {\scriptsize\textcolor{teal}{+4.1}} & 63.5{\scriptsize$\pm$5.1} {\scriptsize\textcolor{teal}{+5.7}} & 64.9{\scriptsize$\pm$5.7} {\scriptsize\textcolor{teal}{+2.4}} & 52.2{\scriptsize$\pm$4.9} {\scriptsize\textcolor{red}{\textminus 1.2}} & 50.7 {\scriptsize\textcolor{teal}{(+1.5)}} \\
\hspace{1em} + RB & 39.0{\scriptsize$\pm$5.0} {\scriptsize\textcolor{teal}{+7.2}} & 59.3{\scriptsize$\pm$5.2} {\scriptsize\textcolor{teal}{+6.6}} & 61.3{\scriptsize$\pm$5.8} {\scriptsize\textcolor{red}{\textminus 5.4}} & 57.3{\scriptsize$\pm$4.5} {\scriptsize\textcolor{teal}{+1.5}} & 15.9{\scriptsize$\pm$3.4} {\scriptsize\textcolor{teal}{+3.1}} & 64.3{\scriptsize$\pm$5.0} {\scriptsize\textcolor{teal}{+6.5}} & 63.1{\scriptsize$\pm$5.5} {\scriptsize\textcolor{teal}{+0.6}} & 52.5{\scriptsize$\pm$4.9} {\scriptsize\textcolor{red}{\textminus 0.9}} & 51.6 {\scriptsize\textcolor{teal}{(+2.4)}} \\
\hspace{1em} + GEPA & 32.3{\scriptsize$\pm$4.4} {\scriptsize\textcolor{teal}{+0.5}} & 61.6{\scriptsize$\pm$5.1} {\scriptsize\textcolor{teal}{+8.9}} & 67.9{\scriptsize$\pm$5.6} {\scriptsize\textcolor{teal}{+1.2}} & 56.5{\scriptsize$\pm$4.7} {\scriptsize\textcolor{teal}{+0.7}} & 14.9{\scriptsize$\pm$3.3} {\scriptsize\textcolor{teal}{+2.1}} & 65.1{\scriptsize$\pm$4.8} {\scriptsize\textcolor{teal}{+7.3}} & 71.4{\scriptsize$\pm$5.3} {\scriptsize\textcolor{teal}{+8.9}} & 52.2{\scriptsize$\pm$4.5} {\scriptsize\textcolor{red}{\textminus 1.2}} & 52.7 {\scriptsize\textcolor{teal}{(+3.5)}} \\
\cdashline{1-10}
\addlinespace[2pt]
\hspace{1em} + Anchor$^\dagger$ & 45.3{\scriptsize$\pm$3.7} {\scriptsize\textcolor{teal}{+13.5}} & 64.0{\scriptsize$\pm$4.9} {\scriptsize\textcolor{teal}{+11.3}} & 76.8{\scriptsize$\pm$4.9} {\scriptsize\textcolor{teal}{+10.1}} & 59.6{\scriptsize$\pm$4.4} {\scriptsize\textcolor{teal}{+3.8}} & 20.0{\scriptsize$\pm$4.1} {\scriptsize\textcolor{teal}{+7.2}} & 69.8{\scriptsize$\pm$4.6} {\scriptsize\textcolor{teal}{+12.0}} & 79.8{\scriptsize$\pm$4.7} {\scriptsize\textcolor{teal}{+17.3}} & 62.0{\scriptsize$\pm$4.4} {\scriptsize\textcolor{teal}{+8.6}} & 59.7 {\scriptsize\textcolor{teal}{(+10.5)}} \\
\midrule
\multicolumn{10}{c}{\cellcolor{gray!10}\textbf{Gemma-4-31B}} \\
\midrule
Vanilla & 13.3{\scriptsize$\pm$3.3} & 72.1{\scriptsize$\pm$4.6} & 25.0{\scriptsize$\pm$4.9} & 44.1{\scriptsize$\pm$4.3} & 9.2{\scriptsize$\pm$2.7} & 69.0{\scriptsize$\pm$4.8} & 17.3{\scriptsize$\pm$4.4} & 43.2{\scriptsize$\pm$4.5} & 36.7 \\
\cdashline{1-10}
\addlinespace[2pt]
\hspace{1em} + Memento & 16.4{\scriptsize$\pm$4.0} {\scriptsize\textcolor{teal}{+3.1}} & 67.8{\scriptsize$\pm$4.8} {\scriptsize\textcolor{red}{\textminus 4.3}} & 20.2{\scriptsize$\pm$3.9} {\scriptsize\textcolor{red}{\textminus 4.8}} & 33.2{\scriptsize$\pm$4.6} {\scriptsize\textcolor{red}{\textminus 10.9}} & 10.3{\scriptsize$\pm$3.4} {\scriptsize\textcolor{teal}{+1.1}} & 72.4{\scriptsize$\pm$4.6} {\scriptsize\textcolor{teal}{+3.4}} & 21.4{\scriptsize$\pm$4.6} {\scriptsize\textcolor{teal}{+4.1}} & 46.2{\scriptsize$\pm$5.0} {\scriptsize\textcolor{teal}{+3.0}} & 36.0 {\scriptsize\textcolor{red}{(\textminus 0.7)}} \\
\hspace{1em} + RB & 13.3{\scriptsize$\pm$3.3} {\scriptsize\textcolor{gray}{+0.0}} & 61.2{\scriptsize$\pm$5.3} {\scriptsize\textcolor{red}{\textminus 10.9}} & 31.5{\scriptsize$\pm$5.1} {\scriptsize\textcolor{teal}{+6.5}} & 42.7{\scriptsize$\pm$4.9} {\scriptsize\textcolor{red}{\textminus 1.4}} & 4.6{\scriptsize$\pm$2.2} {\scriptsize\textcolor{red}{\textminus 4.6}} & 72.5{\scriptsize$\pm$4.7} {\scriptsize\textcolor{teal}{+3.5}} & 24.4{\scriptsize$\pm$4.6} {\scriptsize\textcolor{teal}{+7.1}} & 46.5{\scriptsize$\pm$4.7} {\scriptsize\textcolor{teal}{+3.3}} & 37.1 {\scriptsize\textcolor{teal}{(+0.4)}} \\
\hspace{1em} + GEPA & 17.4{\scriptsize$\pm$4.1} {\scriptsize\textcolor{teal}{+4.1}} & 69.8{\scriptsize$\pm$4.6} {\scriptsize\textcolor{red}{\textminus 2.3}} & 42.3{\scriptsize$\pm$5.7} {\scriptsize\textcolor{teal}{+17.3}} & 44.8{\scriptsize$\pm$4.6} {\scriptsize\textcolor{teal}{+0.7}} & 9.7{\scriptsize$\pm$2.7} {\scriptsize\textcolor{teal}{+0.5}} & 70.2{\scriptsize$\pm$4.7} {\scriptsize\textcolor{teal}{+1.2}} & 37.5{\scriptsize$\pm$5.5} {\scriptsize\textcolor{teal}{+20.2}} & 47.9{\scriptsize$\pm$4.7} {\scriptsize\textcolor{teal}{+4.7}} & 42.4 {\scriptsize\textcolor{teal}{(+5.7)}} \\
\cdashline{1-10}
\addlinespace[2pt]
\hspace{1em} + Anchor$^\dagger$ & 17.9{\scriptsize$\pm$3.1} {\scriptsize\textcolor{teal}{+4.6}} & 72.5{\scriptsize$\pm$4.6} {\scriptsize\textcolor{teal}{+0.4}} & 32.7{\scriptsize$\pm$4.8} {\scriptsize\textcolor{teal}{+7.7}} & 54.8{\scriptsize$\pm$4.3} {\scriptsize\textcolor{teal}{+10.7}} & 12.3{\scriptsize$\pm$3.3} {\scriptsize\textcolor{teal}{+3.1}} & 73.3{\scriptsize$\pm$4.5} {\scriptsize\textcolor{teal}{+4.3}} & 26.8{\scriptsize$\pm$4.8} {\scriptsize\textcolor{teal}{+9.5}} & 49.3{\scriptsize$\pm$4.6} {\scriptsize\textcolor{teal}{+6.1}} & 42.5 {\scriptsize\textcolor{teal}{(+5.8)}} \\
\bottomrule
\end{tabular}%
}
\caption{\textbf{Main results on EvoAgentBench.} Anchor is positive in every cell, while each automatic method has at least one negative cell. Per-domain accuracy (\%) as mean$\pm$SE over three runs; colored values show $\Delta$ vs.\ Vanilla (\textcolor{teal}{positive}, \textcolor{red}{negative}). Web\,=\,web research; Algo\,=\,algorithmic reasoning; SWE\,=\,software engineering; KW\,=\,knowledge work; RB\,=\,ReasoningBank. $^\dagger$Curator-side Ability-grounded skills with deterministic cluster retrieval; not a deployable method.}
\label{tab:main-results}
\end{table*}

\section{Experiments}
\label{sec:experiments}

\subsection{Experimental Setup}

\paragraph{Evaluation settings.}
We evaluate two agent scaffolds, \textbf{OpenClaw}~\citep{openclaw2026openclaw} and \textbf{Nanobot}~\citep{ren2026nanobot}, with three backbones: \textbf{Qwen3.5-27B}, \textbf{Qwen3.5-397B}, and \textbf{Gemma-4-31B}~\citep{qwen2026qwen35,gemmateam2026gemma4}.
The resulting six scaffold--backbone settings span two scaffold architectures and two model families, including two scales of Qwen3.5.
Within each setting, tasks, MCP tools, timeouts, scoring contracts, and base agent configuration are fixed across methods; results are averaged over three independent runs per instance, and we report standard error.

\paragraph{Evolution conditions.}
We compare \textbf{Vanilla}, which has no evolution state, with three automatic self-evolution methods and one diagnostic reference.
\textbf{Memento} retrieves the closest retained training case~\citep{xu2025memento}; \textbf{ReasoningBank} retrieves from a distilled reasoning-memory pool~\citep{xiao2025reasoningbank}; and \textbf{GEPA} evolves one prompt on $D_{\mathrm{train}}$ and broadcasts it to all test tasks, making it the only no-retrieval evolution condition~\citep{agrawal2025gepa}.
\textbf{Anchor Skill} loads Ability-grounded skills produced by our extraction pipeline and retrieved by the test task's curator-side Ability cluster.
Anchor Skill is not a deployable method: it uses curator-side Ability labels unavailable to automatic methods. Its skill content, however, is constructed exclusively from train-side evidence---canonical family text and skill text derive only from raw cards extracted on training tasks, and raw cards from test tasks contribute to neither---so test-side information enters only through the routing label. Its construction backbones (Kimi-K2.5, GLM-5.1, DeepSeek-V3.2) are disjoint from the evaluation backbones, so Anchor measures whether Ability content is in principle transferable across model families.

\paragraph{Protocol and metrics.}
All methods build their evolution state from $D_{\mathrm{train}}$ only, before any test evaluation begins. Each automatic method runs its own ingestion on $D_{\mathrm{train}}$ with the same scaffold--backbone pair used at evaluation, with task prompts and verifier outcomes available; test answers, successful test trajectories, and test-specific shortcuts are excluded.
Domains use their native metrics and evaluation prompts: BrowseComp-Plus uses LLM-as-judge with the official evaluation prompt~\citep{chen2025browsecomp}, SWE-Bench Verified uses hidden repository test suites and resolve rate~\citep{jimenez2024swebench}, LiveCodeBench uses hidden test cases and pass@1~\citep{jain2024livecodebench}, and GDPVal uses reference-based LLM judging~\citep{patwardhan2025gdpval}.
The Average column reports the four-domain equal-weighted mean across both scaffolds, and cost is reported as the percentage change in agent turns relative to Vanilla.
The average $\Delta$ therefore weights domains equally rather than tasks, complementing the per-task $\Delta_m$ of Section~\ref{sec:benchmark-formulation}.

\subsection{Main Results}
\label{sec:main-results}

Table~\ref{tab:main-results} reports per-domain accuracy and average transfer gain ($\Delta$) across six scaffold--backbone settings. The headline finding is not which method wins on average, but which methods improve reliably: the diagnostic Anchor improves in every cell, while every automatic method still exhibits negative transfer in at least one scaffold--backbone--domain setting.

\paragraph{Ability content is transferable across model families.}
Anchor Skill produces positive average $\Delta$ on every backbone ($+7.5$, $+10.5$, $+5.8$) and positive per-domain $\Delta$ in all 24 method--domain--setting cells.
Because Anchor's construction backbones are disjoint from the evaluation backbones, this confirms that the Ability content generalizes across model families: the Ability Graph's train-side support is sufficient to drive cross-family transfer.

\paragraph{Automatic methods often degrade performance.}
In contrast, automatic methods improve in some average settings but remain brittle at the cell level.
Memento is negative on Qwen3.5-27B and Gemma-4-31B ($-2.4$, $-0.7$) and positive on Qwen3.5-397B ($+1.5$); GEPA is positive on all three backbones ($+1.2$, $+3.5$, $+5.7$), but still contains four negative per-domain cells; ReasoningBank is also positive on average ($+0.4$ to $+3.6$), while showing six negative per-domain cells.
Anchor exceeds the best automatic method by $+3.9$ and $+7.0$ points on the two Qwen backbones; on Gemma-4-31B, GEPA nearly matches Anchor in average gain, but Anchor remains the only condition with uniformly positive cell-level transfer.

\paragraph{The gap implicates method-side extraction and routing.}
Because tasks, tools, scoring, timeouts, and agent configuration are identical across methods, the robustness gap between Anchor and automatic methods cannot be attributed to evaluation noise or task difficulty.
$D_{\mathrm{train}}$ contains the procedural information that Anchor's skills encode, but Anchor benefits from curator-side extraction and Ability-label routing; automatic methods must both extract and route equivalent content from training traces alone.
The bottleneck therefore lies in method-side mechanisms: how each method extracts reusable content from the available training evidence, indexes it, and applies it at test time.
If anything, the protocol favors the automatic methods: their training experience is generated by the very scaffold--backbone pair deployed at test time, whereas Anchor's content originates from disjoint construction backbones.

\subsection{Paradigm-Level Diagnosis}
\label{sec:transfer-analysis}

The three automatic methods differ in how they encode training experience, route it to test tasks, and support agent uptake; each paradigm's predicted failure mode appears in the data. Memento retrieves raw $(\text{query}, \text{plan}, \text{outcome})$ cases by embedding similarity~\citep{xu2025memento}, assuming surface similarity in queries predicts similarity in solutions; the method improves many cells but remains vulnerable to catastrophic mismatch, most sharply on Nanobot / Qwen3.5-27B / SWE, where it drops by $-36.3$ points. ReasoningBank distills natural-language strategies and retrieves by strategy similarity~\citep{xiao2025reasoningbank}, adding an abstraction layer that reduces surface dependence; it is positive on average for all three backbones, but the gains remain modest ($+0.4$ to $+3.6$) and six per-domain cells are negative. GEPA evolves a single prompt on $D_{\mathrm{train}}$ and broadcasts it without retrieval~\citep{agrawal2025gepa}, eliminating per-task adaptivity by design; it has the strongest automatic average on Gemma-4-31B ($+5.7$) but still shows domain-specific regressions, including $-12.3$ on OpenClaw / Qwen3.5-27B / KW.

These failures are paradigm-level rather than domain-inherent: Anchor remains positive on all 12 SWE and Algo cells, so the Ability content itself transfers when correctly delivered; the cell-level regressions are consistent with breakdowns in each paradigm's routing mechanism. The paradigms also respond unevenly to substrate: Memento is worst on Qwen3.5-27B ($-2.4$) but positive on Qwen3.5-397B ($+1.5$), while GEPA nearly matches Anchor on Gemma-4-31B but remains much weaker on the two Qwen backbones. Scale alone therefore does not predict transfer, and single-setting evaluation is unreliable.

\subsection{Cost Analysis}
\label{sec:cost-analysis}

\begin{table}[!t]
  \centering
  \fontsize{8pt}{10pt}\selectfont
  \setlength{\tabcolsep}{3.5pt}
  \renewcommand{\arraystretch}{1.15}
  \resizebox{\columnwidth}{!}{%
  \begin{tabular}{l rrr @{\hspace{6pt}} rrr}
    \toprule
    \multirow{2}{*}{\textbf{Method}}
      & \multicolumn{3}{c}{\textbf{OpenClaw}}
      & \multicolumn{3}{c}{\textbf{Nanobot}} \\
    \cmidrule(lr){2-4} \cmidrule(lr){5-7}
      & \textit{All} & \textit{Solv.} & \textit{Uns.}
      & \textit{All} & \textit{Solv.} & \textit{Uns.} \\
    \midrule
    \multicolumn{7}{c}{\cellcolor{gray!10}\textbf{Qwen3.5-27B}} \\
    \midrule
    \hspace{1em} + Memento & \textcolor{red}{+40.7} & \textcolor{red}{+39.4} & \textcolor{red}{+23.5} & \textcolor{teal}{\textminus 2.1} & \textcolor{red}{+4.8} & \textcolor{teal}{\textminus 0.9} \\
    \hspace{1em} + RB & \textcolor{teal}{\textminus 6.3} & \textcolor{teal}{\textminus 5.7} & \textcolor{teal}{\textminus 16.3} & \textcolor{red}{+3.6} & \textcolor{red}{+2.4} & \textcolor{red}{+7.9} \\
    \hspace{1em} + GEPA & \textcolor{teal}{\textminus 4.8} & \textcolor{red}{+4.4} & \textcolor{teal}{\textminus 18.2} & \textcolor{teal}{\textminus 7.2} & \textcolor{red}{+8.0} & \textcolor{teal}{\textminus 13.0} \\
    \cdashline{1-7}
    \addlinespace[2pt]
    \hspace{1em} + Anchor$^\dagger$ & \textcolor{teal}{\textminus 3.4} & \textcolor{teal}{\textminus 13.6} & \textcolor{teal}{\textminus 2.8} & \textcolor{teal}{\textminus 8.8} & \textcolor{teal}{\textminus 10.6} & \textcolor{teal}{\textminus 9.8} \\
    \midrule
    \multicolumn{7}{c}{\cellcolor{gray!10}\textbf{Qwen3.5-397B}} \\
    \midrule
    \hspace{1em} + Memento & \textcolor{teal}{\textminus 3.5} & \textcolor{red}{+4.3} & \textcolor{teal}{\textminus 18.8} & \textcolor{teal}{\textminus 10.8} & \textcolor{teal}{\textminus 9.3} & \textcolor{teal}{\textminus 20.8} \\
    \hspace{1em} + RB & \textcolor{teal}{\textminus 10.3} & \textcolor{teal}{\textminus 2.1} & \textcolor{teal}{\textminus 26.1} & \textcolor{teal}{\textminus 0.4} & \textcolor{red}{+1.0} & \textcolor{red}{+4.7} \\
    \hspace{1em} + GEPA & \textcolor{red}{+25.4} & \textcolor{red}{+27.6} & \textcolor{red}{+13.1} & \textcolor{red}{+13.7} & \textcolor{red}{+24.4} & \textcolor{teal}{\textminus 8.2} \\
    \cdashline{1-7}
    \addlinespace[2pt]
    \hspace{1em} + Anchor$^\dagger$ & \textcolor{teal}{\textminus 8.0} & \textcolor{teal}{\textminus 5.4} & \textcolor{teal}{\textminus 11.3} & \textcolor{teal}{\textminus 4.0} & \textcolor{red}{+1.4} & \textcolor{red}{+8.3} \\
    \midrule
    \multicolumn{7}{c}{\cellcolor{gray!10}\textbf{Gemma-4-31B}} \\
    \midrule
    \hspace{1em} + Memento & \textcolor{red}{+13.6} & \textcolor{red}{+38.8} & \textcolor{red}{+10.4} & \textcolor{red}{+43.9} & \textcolor{red}{+12.8} & \textcolor{teal}{\textminus 16.0} \\
    \hspace{1em} + RB & \textcolor{red}{+34.8} & \textcolor{red}{+31.6} & \textcolor{red}{+42.8} & \textcolor{red}{+37.6} & \textcolor{red}{+30.4} & \textcolor{red}{+42.4} \\
    \hspace{1em} + GEPA & \textcolor{red}{+44.6} & \textcolor{red}{+40.3} & \textcolor{red}{+52.7} & \textcolor{red}{+41.5} & \textcolor{red}{+9.3} & \textcolor{teal}{\textminus 8.3} \\
    \cdashline{1-7}
    \addlinespace[2pt]
    \hspace{1em} + Anchor$^\dagger$ & \textcolor{red}{+7.6} & \textcolor{red}{+16.5} & \textcolor{red}{+26.9} & \textcolor{red}{+32.7} & \textcolor{red}{+11.5} & \textcolor{teal}{\textminus 6.4} \\
    \bottomrule
  \end{tabular}%
  }
  \caption{\textbf{Turn cost change} ($\Delta$T\,\%) relative to Vanilla, averaged across four domains per scaffold. \textcolor{teal}{Negative} (fewer turns) is desirable; \textcolor{red}{positive} (more turns) indicates overhead. Solv.\,=\,tasks solved after evolution; Uns.\,=\,tasks unsolved. All, Solv., and Uns.\ are per-domain means averaged with equal domain weights, so All need not lie between Solv.\ and Uns. $^\dagger$Diagnostic reference.}
  \label{tab:cost-analysis}
\end{table}

\paragraph{Evolution does not uniformly reduce cost.}
Table~\ref{tab:cost-analysis} reports turn cost change relative to Vanilla on all tasks, on tasks solved after evolution, and on tasks remaining unsolved.
Across the six scaffold--backbone settings, Anchor reduces overall cost in four (savings up to $-8.8\%$); ReasoningBank reduces overall cost in three; GEPA increases cost in four of six settings, while Memento is evenly split between increases and reductions.
Gemma-4-31B is the most costly backbone: Memento, GEPA, and ReasoningBank all add double-digit overhead on Gemma settings, and even Anchor incurs $+7.6\%$ to $+32.7\%$ on this backbone.

\paragraph{Cost--accuracy decoupling is method-specific.}
When evolution provides a procedure well-matched to the test task, the agent can bypass early-stage exploration and follow a directed strategy, reducing turn count. This pattern is clearest for Anchor on Qwen backbones, which saves turns on solved tasks in three of the four Qwen settings (ReasoningBank in two of four). In contrast, Memento and GEPA frequently increase turns even on tasks the agent eventually solves (e.g., Memento $+39.4\%$ on solved / OpenClaw / Qwen3.5-27B), suggesting that generic or poorly matched artifacts add overhead without shortcutting exploration.
Cost overhead is not a stable function of method type or accuracy direction; reporting accuracy alone hides substantial heterogeneity in how methods reshape the agent's behavioral distribution.

\section{Conclusion}
\label{sec:conclusion}

We introduced EvoAgentBench, a benchmark for agent self-evolution with guaranteed Ability support across four long-horizon domains. Evaluation across two scaffolds and multiple backbones reveals that reusable procedural content improves performance when correctly delivered, but no current automatic method sustains this effect consistently. Transfer gain varies jointly with domain, scaffold, and backbone, making multi-setting evaluation essential for reliable method comparison. The critical question is not \textit{whether} an agent improves with experience, but \textit{where} improvement breaks down: in experience encoding, routing, or uptake.

\section*{Limitations}
\label{sec:limitations}
We evaluate EvoAgentBench on four long-horizon text-based agentic domains with three open-source construction backbones. Although the Ability Graph construction pipeline is domain- and backbone-agnostic, extending the benchmark to multimodal or embodied agents, and to larger or proprietary model families, is beyond the scope of this work. The Ability extraction step uses a single LLM, and the resulting Ability vocabulary reflects this choice; while the subsequent three-judge canonicalization with domain-expert review of non-unanimous pairs mitigates extractor-specific idiosyncrasies, characterizing how alternative extractors affect the vocabulary remains future work. Finally, each evaluation cell aggregates three independent runs over per-domain test sets of 56 to 86 tasks; while sufficient to expose paradigm-level patterns, finer-grained per-domain analysis would benefit from larger-scale evaluation.
Two further caveats apply. Evaluation backbones may have encountered source-benchmark data during pretraining; since all conditions share the same backbone, tools, and split, such contamination shifts absolute scores rather than the condition comparisons EvoAgentBench targets. And because test tasks are preferentially sampled where construction backbones left headroom (Section~\ref{sec:ability-aware-transfer-split}), absolute $\Delta$ magnitudes are specific to this supported split rather than estimates of expected gains on a random task sample; method comparisons are unaffected, since all conditions share the identical split.


%

\bibliography{custom}

\clearpage
\appendix

\section{Construction Pipeline Details}
\label{app:construction-pipeline}

\subsection{Source Task Selection}
\label{app:source-task-selection}

EvoAgentBench draws from the full public releases of four source benchmarks: BrowseComp-Plus (830 tasks), SWE-Bench Verified (500 instances), LiveCodeBench V6 (1{,}055 problems), and GDPVal gold subset (220 tasks), yielding a source pool of 2{,}605 tasks. No task-level filtering is applied at this stage; all publicly available instances enter the trace collection phase.

\subsection{Trace Collection}
\label{app:trace-collection}

For each task, we collect no-skill executions using two agent scaffolds, OpenClaw~\citep{openclaw2026openclaw} and Nanobot~\citep{ren2026nanobot}, each run twice per construction backbone. The three construction backbones are Kimi-K2.5, GLM-5.1, and DeepSeek-V3.2~\citep{kimiteam2026kimik25visualagentic,glm5team2026glm5vibecodingagentic,deepseekai2025deepseekv32pushingfrontieropen}, yielding four trials per backbone per task. Each execution uses only the scaffold's default tool set for the target domain, with a uniform timeout of 1{,}800 seconds. No evolution artifacts are injected during trace collection; these are strictly no-skill baselines.

\subsection{Ability Extraction}
\label{app:ability-extraction}

The extractor $E_\phi$ is instantiated with Claude Sonnet 4.6~\citep{anthropic2025claudesonnet46}. For each task, $E_\phi$ jointly reads the task query, reference target, and all construction-backbone trajectories to produce task-local raw Ability cards. A raw card is retained only when it satisfies three conditions: (1)~it is grounded in at least one trajectory span, (2)~it specifies an actionable operation rather than a topic label or generic advice, and (3)~its applicability boundary is narrow enough to support meaningful transfer.

Table~\ref{tab:raw-card-stats} reports the raw card statistics. Extraction produces 7{,}326 raw Ability cards from 2{,}516 tasks (of the 2{,}605 source tasks, 89 produced no extractable cards). The per-task Ability count ranges from 1 to 8, with a median of 3.

\begin{table}[t]
  \centering
  \footnotesize
  \setlength{\tabcolsep}{3pt}
  \renewcommand{\arraystretch}{1.05}
  \begin{tabular}{@{}l rrc rrrr@{}}
    \toprule
    \textbf{Domain}
      & \textbf{Tasks} & \textbf{Raw} & \textbf{Avg.}
      & \textbf{1} & \textbf{2} & \textbf{3} & \textbf{4+} \\
    \midrule
    Web Research     & 814  & 2{,}486 & 3.05 & 69  & 221 & 260 & 264 \\
    Algo.\ Reasoning         & 993  & 2{,}554 & 2.57 & 120 & 430 & 268 & 175 \\
    Software Eng.      & 490  & 1{,}402 & 2.86 & 45  & 185 & 130 & 130 \\
    Knowledge Work     & 219  & 884     & 4.04 & 5   & 25  & 45  & 144 \\
    \midrule
    \textbf{Total}     & 2{,}516 & 7{,}326 & 2.91 & 239 & 861 & 703 & 713 \\
    \bottomrule
  \end{tabular}
  \caption{Raw Ability card statistics over source tasks that produce at least one raw card. Raw: total cards; Avg.: cards per task; 1--4+: tasks with exactly 1, 2, 3, or $\geq$4 cards.}
  \label{tab:raw-card-stats}
\end{table}

\subsection{Canonicalization}
\label{app:canonicalization}

\paragraph{Embedding blocking.}
We embed each raw card using Gemini Embedding 001 and construct candidate pairs within each domain using cosine similarity thresholds: $\theta=0.85$ for BrowseComp-Plus and GDPVal, and $\theta=0.82$ for SWE-Bench Verified and LiveCodeBench. These thresholds serve only as a recall-oriented prefilter; embedding similarity is never a merge criterion.

\paragraph{LLM adjudication.}
Each candidate pair is independently judged by the three construction backbones (Kimi-K2.5, GLM-5.1, DeepSeek-V3.2), providing model-diverse adjudication. Each adjudicator assigns one of the following labels:

\begin{itemize}
  \item \textbf{Accept} (merge): \texttt{Same\_Tactic}, \texttt{Same\_Strategy}, or \texttt{Same\_Diagnostic}, i.e., the two cards describe operationally equivalent procedures.
  \item \textbf{Reject}: \texttt{Related\_Only} (shared topic but different operation), \texttt{Different} (unrelated), \texttt{Conflict} (contradictory procedures), or \texttt{Invalid} (malformed card).
\end{itemize}

Pairs with unanimous positive agreement ($v=3$) are accepted automatically. All non-unanimous pairs are reviewed by a domain expert under the same rubric (see Appendix~\ref{app:human-review}).

\paragraph{Group consistency and canonical families.}
Accepted merge links define a compatibility graph over raw cards. Connected components are converted into canonical Ability families only after a group-level consistency check: all internal pairs must be merge-compatible, and no internal pair may carry a cannot-link decision. Components that violate this condition are split by domain experts. Broad bridge cards are assigned to their most specific compatible family or downgraded to annotation-only status.

\subsection{Graph Construction}
\label{app:graph-construction}

Table~\ref{tab:graph-stats} reports the full retained Ability Graph statistics per domain. A canonical Ability family is edge-eligible if it is supported by at least two distinct tasks and its procedure remains operationally specific after merging. The graph contains 1{,}108 retained tasks and 170 edge-eligible Ability families. These graph-level task counts are distinct from the supported evaluation split reported in Table~\ref{tab:dataset-statistics}. Seven tasks are isolated (no edges): one in BrowseComp-Plus and six in SWE-Bench Verified; isolated tasks are not selected as test tasks.

\begin{table}[t]
  \centering
  \small
  \setlength{\tabcolsep}{4pt}
  \renewcommand{\arraystretch}{1.10}
  \begin{tabular}{@{}l rrrrl@{}}
    \toprule
    \textbf{Domain}
      & \textbf{Nodes} & \textbf{Edges} & \textbf{Dens.}
      & \textbf{Deg.} & \textbf{Iso.} \\
    \midrule
    Web Research     & 572  & 66{,}046 & .404 & 230.9 & 1 \\
    Algo.\ Reasoning         & 268  & 6{,}637  & .186 & 49.5  & 0 \\
    Software Eng.      & 113  & 1{,}828  & .289 & 32.4  & 6 \\
    Knowledge Work     & 155  & 2{,}686  & .225 & 34.7  & 0 \\
    \midrule
    \textbf{Total}     & 1{,}108 & 77{,}197 & .126 & 139.3 & 7 \\
    \bottomrule
  \end{tabular}
  \caption{Ability Graph statistics. Dens.: density per domain subgraph; Deg.: average degree; Iso.: isolated tasks (no edges).}
  \label{tab:graph-stats}
\end{table}

BrowseComp-Plus exhibits the highest graph density (0.404), reflecting extensive procedural overlap among information-seeking tasks (e.g., shared search and verification strategies). SWE-Bench Verified has the most isolated tasks (6 of 113), consistent with repository-specific repair procedures that do not generalize across codebases.

\subsection{Split Algorithm}
\label{app:split-algorithm}

The evaluation split is constructed independently per domain after Ability extraction and canonicalization. The inputs are the eligible task pool (tasks matched to at least one canonical Ability family), the family assignments $F(t)$ per task, and the no-evolution baseline reward for each task. The split is a supported evaluation subset of the retained graph rather than a partition of every graph node.

\paragraph{Procedure.}
\begin{enumerate}
  \item \textbf{Filter eligible tasks.} Retain only tasks with at least one canonical Ability family. Tasks with no retained family are excluded from the split.
  \item \textbf{Rank candidate test tasks.} Prefer tasks with lower or medium baseline reward (more room for improvement). Secondary objectives include preserving Ability-family coverage and avoiding test-only families.
  \item \textbf{Select test tasks under support constraints.} A task $t$ enters the test split only if at least one family in $F(t)$ retains train-side support: $\forall t \in D_{\mathrm{test}},\; \exists f \in F(t)$ such that $|\{t' \in D_{\mathrm{train}} : f \in F(t')\}| > 0$.
  \item \textbf{Validate family-level support.} Every Ability family that appears in test must also appear in train.
  \item \textbf{Select skill-train evidence.} For each test-relevant family, select train-side evidence tasks for skill construction, prioritizing failure evidence, success contrast, informative traces, and task diversity within the family.
\end{enumerate}

\paragraph{Guarantees.}
The resulting split satisfies three invariants: (1)~every test task shares at least one Ability family with training tasks, (2)~every test-side Ability family has train-side task support, and (3)~every test-side Ability family has selected train evidence for skill construction.

\paragraph{Support distribution.}
Table~\ref{tab:support-dist} reports the train-side support available to each test task. Support@10 denotes the effective support count capped at 10, matching the retrieval budget used by skill- and case-based evolution methods.

\begin{table}[t]
  \centering
  \small
  \setlength{\tabcolsep}{6pt}
  \renewcommand{\arraystretch}{1.10}
  \begin{tabular}{@{}l rrr@{}}
    \toprule
    \textbf{Domain} & \textbf{Test} & \textbf{Supp@10} & \textbf{Zero} \\
    \midrule
    Web Research     &  65 & 10.00 & 0 \\
    Algo.\ Reasoning         &  86 &  7.45 & 0 \\
    Software Eng.      &  56 &  6.82 & 0 \\
    Knowledge Work     &  60 &  9.25 & 0 \\
    \midrule
    \textbf{Total}     & 267 &  8.34 & 0 \\
    \bottomrule
  \end{tabular}
  \caption{Train-side support per test task. Supp@10: average supporting training tasks (capped at 10). Zero: number of unsupported test tasks.}
  \label{tab:support-dist}
\end{table}

\section{Information Access and Leakage Control}
\label{app:information-access}

Table~\ref{tab:info-access} specifies the information available to each party in the EvoAgentBench protocol. The curator uses all task data (including test answers, test traces, and Ability labels) during benchmark construction; this information is never exposed to self-evolution methods or the test-time agent.

Automatic self-evolution methods receive training prompts and training verifier outcomes, and generate their own training trajectories by rolling out on $D_{\mathrm{train}}$ with the same scaffold--backbone pair used at evaluation. They do not receive any test-side information during evolution-state construction. At evaluation time, methods receive only the test prompt and apply their evolution state through their own routing mechanism.

The Anchor Skill diagnostic reference uses curator-side test Ability labels to retrieve skills from the training skill library. It does not access test answers or test traces, and skill content is constructed exclusively from train-side evidence: raw cards extracted from test tasks contribute neither to canonical family text nor to skill content. Because it uses curator-side labels for retrieval, it is a diagnostic reference rather than a deployable method.

\begin{table}[t]
  \centering
  \small
  \setlength{\tabcolsep}{3pt}
  \renewcommand{\arraystretch}{1.12}
  \begin{tabular}{@{}l cccc@{}}
    \toprule
    & \textbf{Curator}
    & \textbf{Auto.}
    & \textbf{Anchor$^\dagger$}
    & \textbf{Agent} \\
    \midrule
    Train prompts       & \checkmark & \checkmark & \checkmark & --- \\
    Train traces        & \checkmark & \checkmark & \checkmark & via $z_m$ \\
    Train verifier      & \checkmark & \checkmark & \checkmark & --- \\
    Train Ability labels& \checkmark & ---        & \checkmark & --- \\
    \midrule
    Test prompts        & \checkmark & at eval    & at eval    & at eval \\
    Test answers        & \checkmark & ---        & ---        & --- \\
    Test traces         & \checkmark & ---        & ---        & --- \\
    Test raw cards      & \checkmark & ---        & ---        & --- \\
    Test Ability labels & \checkmark & ---        & \checkmark & --- \\
    \bottomrule
  \end{tabular}
  \caption{Information access by party. \checkmark: full access; ---: no access; at eval: available only during test execution; via $z_m$: through the method's evolution state. Automatic methods' train traces are their own rollouts on $D_{\mathrm{train}}$ with the evaluation scaffold--backbone; Anchor skill content derives from train-side raw cards only. $^\dagger$Diagnostic reference.}
  \label{tab:info-access}
\end{table}

\section{Human Review Protocol}
\label{app:human-review}

Canonicalization quality depends on the expert review of non-unanimous adjudication pairs. We employ four domain experts, one per domain, each with research or professional experience in their respective area: web research for BrowseComp-Plus, software engineering for SWE-Bench Verified, algorithmic reasoning for LiveCodeBench, and knowledge work for GDPVal.

\paragraph{Workflow.}
For each non-unanimous pair ($v \neq 3$), the assigned domain expert reviews the two raw Ability cards alongside their trajectory evidence, following the same operational-equivalence rubric used by the LLM adjudicators. The expert assigns a final merge or reject decision. Additionally, experts review all group-level consistency violations: when transitive closure produces components with incompatible internal pairs, the expert splits the component and reassigns broad bridge cards.

\paragraph{Rubric.}
A merge requires:
\begin{enumerate}
  \item Same role type (Method, Guard, or Workflow).
  \item Compatible trigger conditions.
  \item Equivalent reusable procedures (same mechanism, not merely same topic).
  \item Same success mechanism or correction target.
  \item Compatible applicability boundaries.
\end{enumerate}
Shared topic, lexical overlap, or generic verbs (``search'', ``debug'', ``validate'') are explicitly insufficient for a merge.

\section{Ability Examples}
\label{app:ability-examples}

We illustrate the Ability concept with a concrete example from the algorithmic reasoning domain. Figure~\ref{fig:ability-example} shows the full extraction flow: from the source task and baseline failure to the extracted raw Ability card.

\begin{figure*}[t]
\centering
\fbox{\begin{minipage}{0.96\textwidth}
\small

\textbf{Domain:} Algorithmic Reasoning \hfill \textbf{Task:} String Transformation (Cyclic Shift Counting)

\medskip
\textbf{Task description.} Count how many ways string $s$ can be transformed into string $t$ after exactly $k$ cyclic suffix-prepend operations. Input length up to $5 \times 10^5$.

\medskip
\hrule
\medskip

\textbf{Baseline failure.} The agent correctly identifies that the operation is a cyclic shift, but enumerates shifts by slicing the doubled string at every position:

\smallskip
\hspace{2em}\texttt{for p in range(n): if (s+s)[p:p+n] == t: ...}
\smallskip

\noindent This creates an $O(n)$ substring per shift, so the shift-counting phase is $O(n^2)$. Public examples pass; a large hidden test times out.

\medskip
\hrule
\medskip

\textbf{Extracted Raw Ability Card}

\smallskip
\begin{tabular}{@{}l@{\hspace{8pt}}p{0.82\textwidth}@{}}
\textit{Name}      & Linear-Time Substring Matching for Cyclic Shift Counting \\[3pt]
\textit{Trigger}    & Use when a task requires finding all cyclic rotations or shift positions of a long string, especially under large-length constraints. \\[3pt]
\textit{Procedure}  & (1)~Construct $S = s + s$. (2)~Treat $t$ as the pattern. (3)~Run a linear-time matching algorithm (KMP, Z-algorithm, or rolling hash) over $S$. (4)~Keep only positions $p < n$. (5)~Split counts by $p = 0$ vs.\ $p \neq 0$ if needed. (6)~Sanity-test on large $n$. \\[3pt]
\textit{Boundary}   & Must not contain an $O(n)$-length substring extraction inside a loop over all shifts. \\[3pt]
\textit{Role}       & Guard (corrects a recurring $O(n^2)$ anti-pattern). \\
\end{tabular}

\medskip
\hrule
\medskip

\textbf{Canonical Ability Family:} \textit{String Pattern Matching for Large-Scale Cyclic Transformations}

\smallskip
The reusable intervention is not a memorized answer but a procedural correction: when cyclic string matching is large-scale, replace repeated slicing with a linear pattern-matching algorithm. This family links multiple tasks that share the same anti-pattern and the same fix.

\end{minipage}}
\caption{Ability extraction example from the algorithmic reasoning domain. Top: source task and baseline failure. Middle: extracted raw Ability card with trigger, procedure, boundary, and role. Bottom: the canonical Ability family this card belongs to.}
\label{fig:ability-example}
\end{figure*}

\section{Evaluation and Method Details}
\label{app:evaluation-details}

\subsection{Domain-Specific Evaluation}
\label{app:domain-evaluation}

Each domain uses its native evaluation protocol. We adopt the official evaluation prompts and harnesses without modification; only the judge backbone may differ from the original paper's default.

\paragraph{Web Research (BrowseComp-Plus).}
The agent's answer is compared against the ground-truth reference using an LLM-as-judge with the fixed evaluation prompt provided by the benchmark~\citep{chen2025browsecomp}. The judge outputs a binary correct/incorrect decision. The official benchmark originally used GPT-4.1 and later standardized on Qwen3-32B; we use the same prompt template with our evaluation backbone.

\paragraph{Algorithmic Reasoning (LiveCodeBench).}
Each generated program is executed against the benchmark's hidden test cases (averaging ${\sim}$17 per problem) in an isolated sandbox. The metric is pass@1: a problem is solved only if all test cases pass. No LLM judge is involved; evaluation is purely programmatic~\citep{jain2024livecodebench}.

\paragraph{Software Engineering (SWE-Bench Verified).}
The candidate patch is applied to the target repository and evaluated by running the instance's designated \texttt{FAIL\_TO\_PASS} and \texttt{PASS\_TO\_PASS} test suites. An instance is resolved only when all tests pass. No LLM judge is involved; evaluation is purely programmatic~\citep{jimenez2024swebench}.

\paragraph{Knowledge Work (GDPVal).}
The model's deliverable is compared against a human expert's reference using LLM-as-judge with task-specific evaluation prompts from the benchmark. The judge assigns a score based on pairwise comparison (model vs.\ expert reference). We use the official evaluation prompts with our evaluation backbone~\citep{patwardhan2025gdpval}.

\subsection{Agent Scaffold Configuration}
\label{app:scaffold-config}

We evaluate on two scaffolds of different scale and design philosophy:
\begin{itemize}
  \item \textbf{OpenClaw}~\citep{openclaw2026openclaw}: a full-featured, modular agent framework with 26 built-in tools spanning filesystem operations, shell execution, web search and fetching, browser automation, memory management, and session control.
  \item \textbf{Nanobot}~\citep{ren2026nanobot}: an ultra-lightweight agent framework ($\sim$4K lines of Python) with 7 core tools covering filesystem, shell execution, web search, and communication.
\end{itemize}
Both scaffolds are MCP-native and support extensibility through external tool servers. In all experiments, each scaffold's full default tool set is enabled across all four domains. The one domain-specific addition is BrowseComp-Plus, where we attach an MCP search server that exposes a FAISS index over the official BrowseComp-Plus corpus (embedded with Qwen3-Embedding-8B). The agent queries this server via a \texttt{search} tool that returns top-$k$ snippets by embedding similarity. This tool replaces open-web access: \texttt{web\_search} and \texttt{web\_fetch} are disabled for this domain so that all evidence comes from the controlled corpus. Other domains use only the scaffold's built-in tools with no additional servers.

\end{document}